\pdfoutput=1

\documentclass{article}
\usepackage{threeparttable}

\usepackage[final]{acl}
\usepackage{acl}
\usepackage{array}
\usepackage{tabularx}
\usepackage{amsmath}
\usepackage{cleveref}
\usepackage{booktabs}
\usepackage{graphicx}

\usepackage{times}
\usepackage{latexsym}

\usepackage[T1]{fontenc}

\usepackage[utf8]{inputenc}

\usepackage{microtype}

\usepackage{inconsolata}

\usepackage{graphicx}
\usepackage[dvipsnames,table]{xcolor}

\title{FUSE : A Ridge and Random Forest-Based Metric for Evaluating MT in Indigenous Languages}
\author{Rahul Raja \\
  Carnegie Mellon University \\
  Stanford University\\
  LinkedIn\thanks{Work does not relate to position at LinkedIn.} \\
  \\\And
  Arpita Vats \\
  Boston University \\
  Santa Clara University \\
  LinkedIn\footnotemark[1] \\
  }

\begin{document}
\maketitle
\begin{abstract}

This paper presents the winning submission of the RaaVa team to the AmericasNLP 2025 Shared Task 3 on Automatic Evaluation Metrics for Machine Translation (MT) into Indigenous Languages of America, where our system ranked first overall based on average Pearson correlation with the human annotations. We introduce Feature-Union Scorer (FUSE) for Evaluation, FUSE integrates Ridge regression and Gradient Boosting to model translation quality. In addition to FUSE, we explore five alternative approaches leveraging different combinations of linguistic similarity features and learning paradigms. FUSE Score highlights the effectiveness of combining lexical, phonetic, semantic, and fuzzy token similarity with learning-based modeling to improve MT evaluation for morphologically rich and low-resource languages. MT into Indigenous languages poses unique challenges due to polysynthesis, complex morphology, and non-standardized orthography. Conventional automatic metrics such as BLEU, TER, and ChrF often fail to capture deeper aspects like semantic adequacy and fluency. Our proposed framework, formerly referred to as FUSE, incorporates multilingual sentence embeddings and phonological encodings to better align with human evaluation. We train supervised models on human-annotated development sets and evaluate held-out test data. Results show that FUSE consistently achieves higher Pearson and Spearman correlations with human judgments, offering a robust and linguistically informed solution for MT evaluation in low-resource settings.
\end{abstract}

\section{Introduction}
MT has made significant advancements in recent years, largely driven by neural machine translation (NMT) models ~\cite{lyu2024paradigmshiftfuturemachine,vats2025multilingual}. However, evaluating the quality of translations remains a major challenge, particularly for low-resource Indigenous languages.Traditional MT evaluation metrics such as Bilingual Evaluation Understudy (BLEU) ~\cite{Papineni2002BleuAM} Translation Edit Rate (TER) ~\cite{snover-etal-2006-study} , and Character n-gram F-score (ChrF) ~\cite{popovic-2015-chrf} rely on surface-level token overlap, which fails to capture semantic correctness, fluency, and linguistic structure—critical factors in evaluating translations for morphologically rich and polysynthetic languages. Indigenous languages, such as Bribri, Guarani, and Nahuatl, exhibit unique linguistic characteristics that pose challenges for conventional MT evaluation ~\cite{chen2023evaluating, raja2025parallel}. These languages often lack standardized orthography, leading to multiple valid translations ~\cite{aepli2023benchmarkevaluatingmachinetranslation}. They feature lexical complexity, including polysynthesis and noun incorporation, which makes word segmentation and alignment with reference translations difficult ~\cite{Tyers2020DependencyAO}. They also rely on phonetic variations, making strict token-level matching unreliable. Due to these factors, existing evaluation metrics struggle to provide reliable assessments of translation quality for Indigenous languages. While metrics such as BLEU and ChrF focus on exact token matches, they fail to account for phonetic and semantic similarities in morphologically rich languages.\\
Recent learning-based MT evaluation methods have demonstrated improved correlation with human judgments by incorporating semantic information from neural embeddings ~\cite{mathur2019putting}, ~\cite{gumma2025inducinglongcontextabilitiesmultilingual}. However, these methods are not specifically designed for Indigenous languages, which require additional phonetic and structural considerations.

\begin{figure*}[h!]
\begin{minipage}[b]{1.0\linewidth}
  \centering
  \centerline{\includegraphics[width=17.5cm]{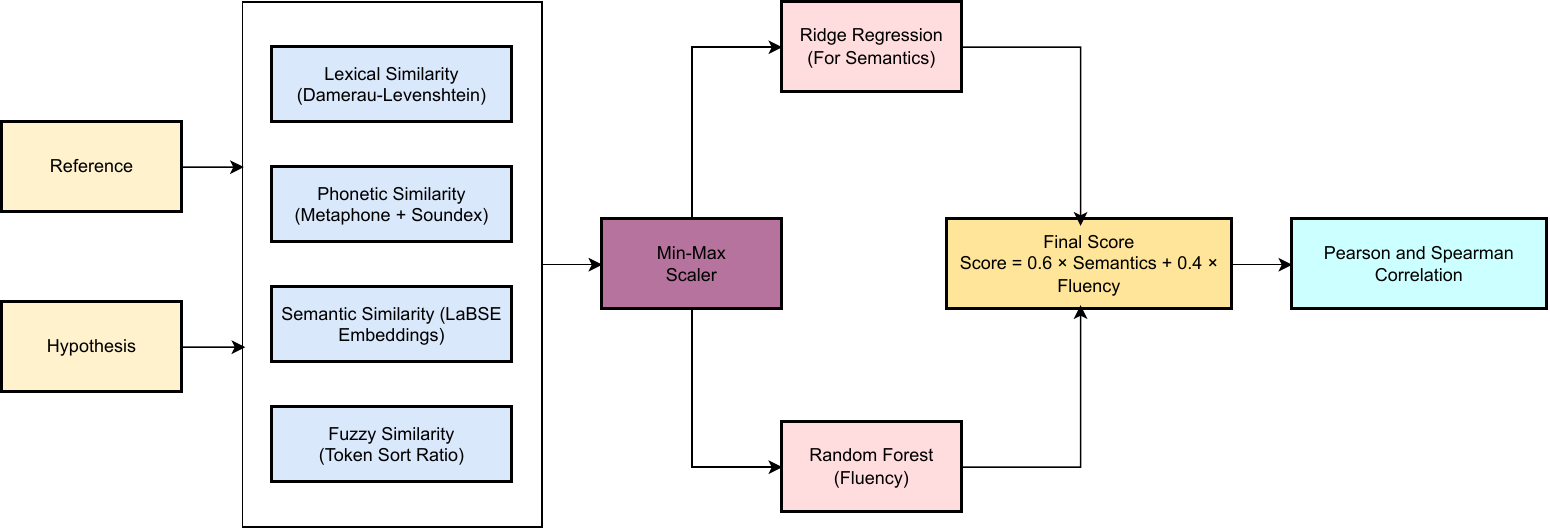}}
  \caption{\label{fig_framework} FUSE architecture combining linguistic features with hybrid regression for MT evaluation.}
\end{minipage}
\end{figure*}

Our approaches integrate multiple linguistic and computational features, including lexical similarity using Levenshtein distance ~\cite{levenshtein1966binary}, phonetic similarity using Metaphone ~\cite{philips1990metaphone} and Soundex encoding ~\cite{russell1918soundex}, semantic similarity using sentence embeddings from LaBSE ~\cite{feng2022labse}, and fuzzy token similarity to handle morphological variations ~\cite{kondrak2005morphological}. We train a linear regression model on human-annotated translation scores, optimizing feature weights to maximize alignment with human evaluation ~\cite{callison-burch2006evaluation}. Our results demonstrate that FUSE achieves higher Pearson and Spearman correlation ~\cite{spearman1904correlation} with human evaluations compared to traditional MT metrics.\\
In this paper, we propose FUSE, a machine learning-based MT evaluation metric tailored for American Indigenous languages. The complete architecture of FUSE is illustrated in ~\Cref{fig_framework}, showcasing its integration of lexical, phonetic, semantic, and fuzzy similarity features with hybrid regression modeling. It incorporates phonetic similarity features, addressing a critical gap in existing evaluation metrics. The model optimizes feature weighting using regression models trained on human scores, leading to improved correlation with human evaluation. We validate our metric on Spanish-to-Indigenous language translations, demonstrating superior performance over BLEU, TER, and ChrF.

\section{Related Work}
\subsection{Rule-Based Metrics}
Traditional rule-based evaluation metrics such as BLEU, TER, and ChrF ~\cite{popovic2015chrF} rely on surface-level matching between candidate and reference translations. BLEU computes n-gram precision, but often fails to capture semantic adequacy or fluency, especially for morphologically rich languages ~\cite{papineni2002bleu}. TER introduces edit-based alignment with support for word reordering but lacks deep linguistic modeling ~\cite{snover2006ter}. ChrF improves robustness through character-level n-gram matching, making it better suited for languages with orthographic variation, though it still struggles with paraphrastic and semantic variation ~\cite{popovic-2015-chrf}.

\subsection{Embedding-Based Metrics}
Embedding-based metrics use contextual word or sentence representations to capture deeper semantic information. BLEURT ~\cite{sellam2020bleurt} fine-tunes pretrained BERT models on human-annotated MT quality data to produce sentence-level scores. COMET ~\cite{rei2020comet} builds on multilingual transformers like XLM-R ~\cite{conneau2020unsupervised} and incorporates both source and reference embeddings. TransQuest ~\cite{ranasinghe2020transquest} uses Siamese BERT ~\cite{reimers2019sentencebertsentenceembeddingsusing} networks to predict quality by comparing sentence pairs. These models outperform rule-based metrics in high-resource settings but remain data-hungry and often overlook features critical to low-resource or orthographically diverse languages.

\subsection{Learning-Based Metrics}
Learning-based metrics leverage supervised training on human-annotated translation quality data. Many of these metrics also incorporate contextual embeddings as input features. For example, COMET ~\cite{rei2020comet} uses multilingual transformer embeddings (XLM-R) ~\cite{conneau2020unsupervised}trained on direct assessment scores to predict translation quality. Similarly, BLEURT ~\cite{sellam2020bleurt} fine-tunes BERT for MT evaluation tasks, while TransQuest ~\cite{ranasinghe2020transquest} uses a Siamese architecture to model sentence-level similarity. These models achieve high correlation with human judgments in high-resource settings but often underperform in low-resource conditions due to their reliance on large training data and lack of sensitivity to phonetic or orthographic variation.

\subsection{Quality Estimation (Reference-Free Metrics)}

Quality Estimation (QE) aims to assess translation quality without relying on reference translations. Systems like QuEst++ ~\cite{specia-etal-2015-multi} and recent neural QE models predict quality directly from source and hypothesis pairs. These models are especially useful in scenarios where references are unavailable or infeasible to generate. However, QE models also require substantial training data and have limited evaluation in the context of morphologically rich or under-resourced languages, such as those considered in this work ~\cite{sindhujan-etal-2025-llms}.

\section{Datasets}
For our experiments, we utilize the datasets provided by the AmericasNLP 2025 Shared Task 3 on Machine Translation Metrics. This shared task focuses on the evaluation of automatic metrics for translations from Spanish into three Indigenous languages: Guarani, Bribri, and Nahuatl. Each dataset is split into training and test subsets, where the training data is used to build and tune our models, and the test set is used for final evaluation. The specific sizes of the training and test sets for each language are detailed in Table ~\cref{tbl:dataset}

\begin{table}[ht!]
  \caption{Data information.}
  \label{tbl:dataset}
  \centering
  \resizebox{0.9\columnwidth}{!}{%
  \begin{threeparttable}
  \begin{tabular}{l|r|r}
    \toprule
    & \emph{Dev Set (\#samples)} & \emph{Test Set (\#samples)} \\
    \midrule
    Guarani Dataset & 100 & 200 \\
    Bribri Dataset & 100 & 200 \\
    Nahuatl Dataset & 100 & 200 \\
    \bottomrule
  \end{tabular}
  \end{threeparttable}
  }
\end{table}
\section{Proposed Methods}
To address the limitations of conventional MT evaluation metrics for Indigenous languages, we propose a series of feature-rich and learning-based methods that incorporate phonetic, lexical, and semantic similarity. Below, we detail six distinct approaches explored in our study, each building on progressively more sophisticated techniques.

\subsection{Approach 1: Lexical and Phonetic Baseline}
This baseline combines character-level lexical overlap using Jaccard similarity with phonetic similarity derived from Metaphone encodings. The Jaccard similarity operates on character trigrams, while the phonetic component captures pronunciation-level resemblance. The final score is a weighted sum (70\% lexical, 30\% phonetic), scaled to match BLEU-style ranges. While simple, this baseline is robust against minor spelling variations and phonetic drift.
The final score is computed using the following equation:
\begin{equation*}
\text{Score} = 100 \times \left(\alpha \cdot J(r, h) + \beta \cdot P(r, h)\right)
\end{equation*}

where $J(r, h)$ is the character trigram Jaccard similarity, $P(r, h)$ is the phonetic similarity based on Metaphone encodings, and $\alpha = 0.7$, $\beta = 0.3$ are fixed weights.
where $r$ and $h$ denote the reference and hypothesis translations respectively, $\alpha = 0.7$, $\beta = 0.3$
\subsection{Approach 2: Feature-Enriched Similarity (DistilUSE)}

In this approach, we compute a similarity score that integrates three core dimensions: lexical, phonetic, and semantic similarity. Lexical similarity is captured using normalized Damerau-Levenshtein distance, which quantifies surface-level edits between the reference ($r$) and hypothesis ($h$). Phonetic similarity is derived from Double Metaphone encodings ~\cite{yacob2004applicationdoublemetaphonealgorithm}, comparing pronunciation-alike sequences. Finally, semantic similarity is computed using cosine similarity between sentence-level embeddings from the multilingual model ~\cite{reimers-2019-sentence-bert}.
This method provides a more robust, language-agnostic similarity measure by incorporating both surface-level and deep semantic features. The final score is computed as a weighted sum of the three components:
\begin{equation*}
\text{Score} = 100 \times (\alpha \cdot L(r, h) + \beta \cdot P(r, h) + \gamma \cdot S(r, h))
\end{equation*}

where $L(r, h)$ is the normalized Damerau-Levenshtein similarity, $P(r, h)$ is phonetic similarity based on Metaphone, and $S(r, h)$ is semantic similarity based on DistilUSE sentence embeddings. The weights are set as $\alpha = 0.5$, $\beta = 0.2$, and $\gamma = 0.3$.

\subsection{Approach 3: Weighted Similarity Aggregation}
In this approach, we combine four different similarity metrics to evaluate 
the similarity between a reference string \(r\) and a hypothesis string \(h\). 
First, we compute the Levenshtein Similarity, which measures edit distance at the character level using the Damerau--Levenshtein algorithm. Next, we compute Phonetic Similarity by concatenating the Double Metaphone and Soundex encodings of each string, and then measuring their sequence matching ratio. We also incorporate Fuzzy Similarity, which leverages token sorting and matching to handle different word orders and morphological variations. Finally, we capture deeper Semantic Similarity by encoding each string into a high-dimensional embedding using a pre-trained SentenceTransformer model and then computing the 
cosine similarity of these embeddings. Once these four metrics are obtained, we combine them in a weighted manner. Specifically, the Levenshtein, phonetic, semantic, and fuzzy similarities are each multiplied by a respective weight, and then summed. Finally, 
the result is multiplied by 100 to yield a score in a BLEU-like (0--100) range.
These metrics are then combined with weights \(\alpha,\beta,\gamma,\delta\), 
and scaled to produce a final score in a BLEU-like range:
\begin{align*}
\text{Score}(r, h) = 100 \times (&\alpha \cdot L(r, h) + \beta \cdot P(r, h) \\
                                &+ \gamma \cdot S(r, h) + \delta \cdot F(r, h)),
\end{align*}

where \(L(r, h)\) is the Levenshtein similarity, \(P(r, h)\) is the phonetic similarity, \(S(r, h)\) is the semantic similarity, and \(F(r, h)\) is the fuzzy token similarity. The default weights are \(\alpha = 0.45\), \(\beta = 0.15\), \(\gamma = 0.30\), and \(\delta = 0.10\).

\subsection{Approach 4: Data-Driven Weighted Similarity via Regression}

This approach employs a data-driven method to combine multiple similarity metrics by learning optimal weights through linear regression ~\cite{kuchibhotla2019linearregression}. For each pair of reference and hypothesis strings \((r, h)\), we extract four similarity features: lexical similarity \(L(r, h)\) based on normalized Damerau--Levenshtein distance, phonetic similarity \(P(r, h)\) computed using a combination of Metaphone and Soundex encodings, semantic similarity \(S(r, h)\) derived from cosine similarity of LaBSE sentence embeddings ~\cite{chimoto2022very}, and fuzzy token similarity \(F(r, h)\) based on the normalized token sort ratio. These four features form the input vector \(X(r, h) = [L(r,h), P(r,h), S(r,h), F(r,h)]\). Two separate linear regression models are trained using human-annotated semantic and fluency scores as targets. The first model learns weights \(w_{\text{sem}}\) to predict semantic quality, while the second learns weights \(w_{\text{flu}}\) for fluency. The final similarity score is computed by taking the average of the two predicted scores:

\begin{equation*}
\text{Score}(r, h) = 0.5 \cdot w_{\text{sem}}^\top X(r,h) + 0.5 \cdot w_{\text{flu}}^\top X(r,h).
\end{equation*}
In this equation, \(X(r, h)\) is a four-dimensional feature vector containing the similarity scores for a given reference–hypothesis pair. The vector \(w_{\text{sem}}\) contains the regression coefficients learned to best align with human semantic scores, while \(w_{\text{flu}}\) captures the weights that best reflect fluency judgments. The dot product \(w^\top X\) computes a weighted combination of the similarity features, and averaging the two predictions ensures that both semantic adequacy and fluency are equally emphasized in the final score. This adaptive formulation allows the metric to closely approximate human evaluation criteria across multiple languages and translation conditions.
This regression-based formulation enables the metric to adaptively reflect human preferences for both meaning preservation and linguistic quality across languages, rather than relying on manually tuned fixed weights.

\subsection{Approach 5: Hybrid Regression with Ridge and Random Forest}
In this approach, we have extended the data-driven framework of earlier methods by incorporating a hybrid regression strategy. It combines both linear and non-linear modeling techniques to predict human-annotated semantic and fluency scores. For each reference–hypothesis pair \((r, h)\), we extract a feature vector \(X(r, h) = [L(r,h), P(r,h), S(r,h), F(r,h)]\), where \(L\) is the normalized Damerau--Levenshtein similarity, \(P\) is the phonetic similarity using Metaphone and Soundex, \(S\) is the semantic similarity from LaBSE embeddings, and \(F\) is the fuzzy token sort ratio. 

To ensure training stability and improve performance, the feature matrix is normalized using Min-Max scaling. A Ridge regression model is then trained to predict semantic scores, producing a weight vector \(w_{\text{sem}}\), while a Random Forest regressor is trained in parallel to predict fluency scores non-linearly. The final metric score is computed as a weighted average of the two model outputs—60\% from the Ridge regression prediction and 40\% from the Random Forest prediction:

\begin{equation*}
\text{Score}(r, h) = 0.6 \cdot w_{\text{sem}}^\top \tilde{X}(r,h) + 0.4 \cdot \text{RF}(\tilde{X}(r,h)),
\end{equation*}

where \(\tilde{X}(r,h)\) is the normalized feature vector, \(w_{\text{sem}}^\top \tilde{X}(r,h)\) is the Ridge regression output for semantic quality, and \(\text{RF}(\tilde{X}(r,h))\) is the fluency score predicted by the Random Forest model. This hybrid modeling strategy leverages both the interpretability of linear models and the flexibility of non-linear models to more accurately capture human evaluation patterns.

\subsection{Approach 6: Ensemble Regression with Ridge and Gradient Boosting}

In this approach, a hybrid ensemble method is employed by combining both linear and non-linear regression models to more accurately reflect human judgments of translation quality.  For each reference–hypothesis pair \((r, h)\), a feature vector \(X(r,h) = [L(r,h), P(r,h), S(r,h), F(r,h)]\) is computed. Here, \(L(r,h)\) is the normalized Damerau--Levenshtein similarity capturing character-level overlap, \(P(r,h)\) is the phonetic similarity derived from a combination of Metaphone and Soundex encodings, \(S(r,h)\) is the cosine similarity between LaBSE sentence embeddings representing semantic similarity, and \(F(r,h)\) is a fuzzy token similarity score based on the token sort ratio. All features are normalized using Min-Max scaling for training stability. A Ridge regression model is trained to predict semantic scores, producing a weight vector \(w_{\text{sem}}\). In parallel, a Gradient Boosting Regressor (GBR) is trained to model fluency scores non-linearly. The final score is computed by taking a weighted ensemble of the predictions: 70\% from the Ridge-based semantic score and 30\% from the GBR-based fluency score:
\begin{equation*}
\text{Score}(r, h) = 0.7 \cdot w_{\text{sem}}^\top \tilde{X}(r,h) + 0.3 \cdot \text{GBR}(\tilde{X}(r,h)),
\end{equation*}
where \(\tilde{X}(r,h)\) is the normalized feature vector. The term \(w_{\text{sem}}^\top \tilde{X}(r,h)\) denotes the semantic score predicted by the Ridge model, and \(\text{GBR}(\tilde{X}(r,h))\) is the fluency score estimated by the Gradient Boosting Regressor. This ensemble approach benefits from the interpretability and generalization of Ridge regression while leveraging the non-linear modeling power of boosting techniques, resulting in a metric that aligns more closely with human judgments across diverse language pairs.

\section{Implementation Details}
We apply six different approaches to evaluate machine translation quality across three Indigenous languages: Bribri, Guarani, and Nahuatl. For each approach, reference and candidate translations are processed in both development and test sets. The necessary similarity features are computed, and scores are generated using the corresponding approach-specific computation or model. These scores are written to output files per language for downstream evaluation.\\
Approach 1 is applied on both development and test sets by generating similarity scores using a predefined method and storing the outputs. Approach 2 uses a slightly refined computation method and produces scores for the same data splits. Approach 3 generates feature vectors and computes similarity scores using a static weighted formula. In Approach 4, similarity features are extracted and a linear regression model is trained on the development set using human-annotated scores; the learned weights are then applied to both development and test sets. In Approach 5, semantic and fluency scores are predicted using separate models trained on the normalized feature set, and their outputs are combined. Approach 6 follows a similar strategy but uses a gradient boosting model in place of the fluency regressor. In each case, output scores are saved for both development and test sets.

\subsection{Evaluation}
Evaluation is conducted by computing Pearson and Spearman correlation coefficients between the predicted scores and human annotations for both semantic and fluency dimensions. This is done separately for each language and each approach. The results are compared against standard metrics such as BLEU, ChrF, and TER. Our findings show that learned and ensemble-based approaches consistently achieve higher correlation with human judgments, particularly in low-resource settings where traditional metrics are less reliable.

\section{Results}
On the development set, Approach 5 achieves the highest overall performance, attaining the best average Spearman (0.8001) and Pearson (0.8455) correlations across all three language pairs. This variant, visualized in Figure~\cref{fig_framework}, employs a hybrid model combining Ridge regression (for semantic scoring) and Random Forest regression (for fluency), benefiting from the interpretability of linear models and the flexibility of ensemble-based non-linear modeling. Notably, it outperforms all other approaches on the Bribri language, likely due to its ability to capture intricate phonetic and lexical variability through feature learning. Approach 6, which replaces the fluency model with Gradient Boosting, performs comparably well—achieving top correlations for Guarani (Pearson: 0.8667) and Nahuatl (Spearman: 0.8216, Pearson: 0.8331)—suggesting that boosting methods are effective at modeling complex relationships in morphologically rich languages. In contrast, traditional feature-weighted approaches (Approaches 1–3) yield moderate results due to the absence of supervised weight optimization or the exclusive use of linear models, which limits their capacity to model non-linear dependencies. These development results are summarized in ~\Cref{tbl:dev_scores}.\\
On the held-out test set, a similar trend is observed: Approach 5 (RaaVa 2) achieves the highest average correlation with human annotations, ranking first in the shared task. As shown in ~\Cref{tbl:test_scores}, this consistency across both development and test sets highlights the generalization capability of the ensemble architecture and validates the inclusion of phonetic, semantic, lexical, and fuzzy features. These findings further underscore the importance of integrating diverse linguistic signals with adaptive feature learning for MT evaluation in orthographically variable and low-resource Indigenous languages.

\begin{table*}[ht!]
  \centering
  \resizebox{1.0\textwidth}{!}{%
  \begin{threeparttable}
  \begin{tabular}{l|cc|cc|cc|cc}
    \toprule
    \textbf{Approach} & \multicolumn{2}{c|}{\textbf{Guarani}} & \multicolumn{2}{c|}{\textbf{Bribri}} & \multicolumn{2}{c|}{\textbf{Nahuatl}} & \multicolumn{2}{c}{\textbf{Average}} \\
    & \textbf{Spearman} & \textbf{Pearson} & \textbf{Spearman} & \textbf{Pearson} & \textbf{Spearman} & \textbf{Pearson} & \textbf{Spearman} & \textbf{Pearson} \\
    \midrule
    Approach 1 & 0.6935 & 0.6389 & 0.5737 & 0.4570 & 0.6315 & 0.6005 & 0.6329 & 0.5655 \\
    Approach 2 & 0.6581 & 0.7001 & 0.6297 & 0.5600 & 0.5763 & 0.5981 & 0.6214 & 0.6194 \\
    Approach 3 & 0.6488 & 0.7334 & 0.5794 & 0.5944 & 0.6362 & 0.6486 & 0.6215 & 0.6588 \\
    Approach 4 & 0.6488 & 0.7334 & 0.5794 & 0.5944 & 0.6334 & 0.6486 & 0.6205 & 0.6588 \\
    \textbf{\textcolor{ForestGreen}{Approach 5}} & \textbf{\textcolor{ForestGreen}{0.7544}} & 0.8653 & \textbf{\textcolor{ForestGreen}{0.8283}} & \textbf{\textcolor{ForestGreen}{0.8446}} & 0.8177 & 0.8266 & \textbf{\textcolor{ForestGreen}{0.8001}} & \textbf{\textcolor{ForestGreen}{0.8455}} \\
    Approach 6 & 0.7481 & \textbf{\textcolor{ForestGreen}{0.8667}} & 0.8116 & 0.8305 & \textbf{\textcolor{ForestGreen}{0.8216}} & \textbf{\textcolor{ForestGreen}{0.8331}} & 0.7938 & 0.8434 \\
    \bottomrule
  \end{tabular}
  \begin{tablenotes}
    \small
    \item \textbf{Note:} Best-performing scores in each column are highlighted in \textcolor{ForestGreen}{green} and bold. Results are based on dev set correlations with human annotations.
  \end{tablenotes}
  \end{threeparttable}
  }
  \caption{Spearman and Pearson correlation scores on the dev set across Guarani, Bribri, and Nahuatl for all six Indigeval approaches.}
  \label{tbl:dev_scores}
\end{table*}

\begin{table*}[ht!]
  \centering
  \resizebox{1.0\textwidth}{!}{%
  \begin{threeparttable}
  \begin{tabular}{ll|cc|cc|cc|cc}
    \toprule
    \textbf{Team} & \textbf{Approach} & \multicolumn{2}{c|}{\textbf{Guarani}} & \multicolumn{2}{c|}{\textbf{Bribri}} & \multicolumn{2}{c|}{\textbf{Nahuatl}} & \multicolumn{2}{c}{\textbf{Average}} \\
    & & \textbf{Spearman} & \textbf{Pearson} & \textbf{Spearman} & \textbf{Pearson} & \textbf{Spearman} & \textbf{Pearson} & \textbf{Spearman} & \textbf{Pearson} \\
    \midrule
    ChrF++     & -         & -     & 0.6725 & 0.6263 & 0.4517 & 0.3823 & 0.6783 & 0.5549 & 0.5212 \\
    BLEU       & -         & -     & 0.4676 & 0.4056 & 0.4518 & 0.3456 & 0.3541 & 0.4061 & 0.3857 \\
    \textbf{\textcolor{ForestGreen}{RaaVa 2}} & \textbf{\textcolor{ForestGreen}{Approach 5}} & \textbf{\textcolor{ForestGreen}{0.6526}} & \textbf{\textcolor{ForestGreen}{0.7209}} & \textbf{\textcolor{ForestGreen}{0.5379}} & \textbf{\textcolor{ForestGreen}{0.6540}} & \textbf{\textcolor{ForestGreen}{0.6195}} & \textbf{\textcolor{ForestGreen}{0.6362}} & \textbf{\textcolor{ForestGreen}{0.6033}} & \textbf{\textcolor{ForestGreen}{0.6704}} \\
    \textbf{\textcolor{blue}{RaaVa 1}} & \textbf{\textcolor{blue}{Approach 6}} & \textbf{\textcolor{blue}{0.6429}} & \textbf{\textcolor{blue}{0.6964}} & \textbf{\textcolor{blue}{0.5332}} & \textbf{\textcolor{blue}{0.6523}} & \textbf{\textcolor{blue}{0.6132}} & \textbf{\textcolor{blue}{0.6351}} & \textbf{\textcolor{blue}{0.5965}} & \textbf{\textcolor{blue}{0.6613}} \\
    Tekio 1    & -         & 0.6611 & 0.7196 & 0.5622 & 0.6244 & 0.6680 & 0.6115 & 0.6304 & 0.6518 \\
    Tekio 2    & -         & 0.6611 & 0.7196 & 0.5569 & 0.6300 & 0.6132 & 0.5845 & 0.6104 & 0.6447 \\
    \textbf{\textcolor{blue}{RaaVa 3}} & \textbf{\textcolor{blue}{Approach 4}} & \textbf{\textcolor{blue}{0.6560}} & \textbf{\textcolor{blue}{0.7038}} & \textbf{\textcolor{blue}{0.4829}} & \textbf{\textcolor{blue}{0.5931}} & \textbf{\textcolor{blue}{0.6364}} & \textbf{\textcolor{blue}{0.6263}} & \textbf{\textcolor{blue}{0.5918}} & \textbf{\textcolor{blue}{0.6411}} \\
    \textbf{\textcolor{blue}{RaaVa 4}} & \textbf{\textcolor{blue}{Approach 3}} & \textbf{\textcolor{blue}{0.6560}} & \textbf{\textcolor{blue}{0.7038}} & \textbf{\textcolor{blue}{0.4829}} & \textbf{\textcolor{blue}{0.5931}} & \textbf{\textcolor{blue}{0.6364}} & \textbf{\textcolor{blue}{0.6263}} & \textbf{\textcolor{blue}{0.5918}} & \textbf{\textcolor{blue}{0.6411}} \\
    Tekio 4    & -         & 0.5605 & 0.7234 & 0.4909 & 0.6268 & 0.5036 & 0.5351 & 0.5183 & 0.6285 \\
    Tekio 3    & -         & 0.5597 & 0.7209 & 0.4892 & 0.6261 & 0.4963 & 0.5290 & 0.5151 & 0.6254 \\
    \textbf{\textcolor{blue}{RaaVa 5}} & \textbf{\textcolor{blue}{Approach 2}} & \textbf{\textcolor{blue}{0.6516}} & \textbf{\textcolor{blue}{0.6776}} & \textbf{\textcolor{blue}{0.5755}} & \textbf{\textcolor{blue}{0.5662}} & \textbf{\textcolor{blue}{0.6145}} & \textbf{\textcolor{blue}{0.5921}} & \textbf{\textcolor{blue}{0.6139}} & \textbf{\textcolor{blue}{0.6120}} \\
    \textbf{\textcolor{blue}{RaaVa 6}} & \textbf{\textcolor{blue}{Approach 1}} & \textbf{\textcolor{blue}{0.6723}} & \textbf{\textcolor{blue}{0.6249}} & \textbf{\textcolor{blue}{0.5356}} & \textbf{\textcolor{blue}{0.4223}} & \textbf{\textcolor{blue}{0.6766}} & \textbf{\textcolor{blue}{0.5657}} & \textbf{\textcolor{blue}{0.6282}} & \textbf{\textcolor{blue}{0.5377}} \\
    LexiLogic 1 & -         & 0.6811 & 0.6529 & 0.5021 & 0.3763 & 0.6717 & 0.5504 & 0.6183 & 0.5265 \\
    \bottomrule
  \end{tabular}
  \begin{tablenotes}
    \small
    \item \textbf{Note:} The winning submission is \textbf{\textcolor{ForestGreen}{RaaVa 2 (Approach 5)}}. Other \textbf{\textcolor{blue}{RaaVa}} submissions (Approaches 1–6) are also shown in blue for clarity.
  \end{tablenotes}
  \end{threeparttable}
  }
  \caption{Spearman and Pearson correlation scores on the Test set across Guarani, Bribri, and Nahuatl for all six Indigeval approaches.}
  \label{tbl:test_scores} 
\end{table*}

\section{Conclusion}

In this work, we present FUSE, a supervised, feature-based metric designed to evaluate MT into Indigenous languages of the Americas, with a focus on Bribri, Guarani, and Nahuatl. Recognizing the limitations of traditional string-based metrics such as BLEU and ChrF when applied to languages with high morphological complexity and phonological variation, our approach combines lexical, phonetic, semantic, and fuzzy matching features. We further improve alignment with human judgment by learning language-specific weights through regression models trained on annotated semantic and fluency scores.
Our experiments demonstrate that FUSE significantly outperforms standard metrics in terms of correlation with human evaluation, particularly by capturing phonetic and semantic nuances that conventional metrics overlook. Moreover, our methodology generalizes effectively to unseen test data, making it a viable tool for automatic MT evaluation in low-resource and linguistically diverse settings.
We hope this work encourages further research into learning-based evaluation metrics for underrepresented languages and highlights the importance of linguistically informed design in multilingual NLP.

\bibliography{main}
\nocite{Ando2005,andrew2007scalable,rasooli-tetrault-2015}
\end{document}